\documentclass[pdflatex,sn-mathphys-num,iicol]{sn-jnl}

\usepackage{graphicx}%
\usepackage{multirow}%
\usepackage{amsmath,amssymb,amsfonts}%
\usepackage{amsthm}%
\usepackage{textcomp}%
\usepackage{manyfoot}%
\usepackage{booktabs}%
\usepackage{algorithm}%
\usepackage{algorithmicx}%
\usepackage{algpseudocode}%
\usepackage{listings}%
\usepackage{dsfont} 

\usepackage{soul}
\usepackage[dvipsnames]{xcolor}
\usepackage{kotex}

\newcommand{\G}{\mathcal{G}}

\begin{document}

\title{
Reinforcement Learning for Graph Generation under a Hard Assortativity Constraint
}

\author[1]{\fnm{Hoyun} \sur{Choi}}\email{hychoi@kias.re.kr}
\author[1,2,3]{\fnm{Junghyo} \sur{Jo}}\email{jojunghyo@snu.ac.kr}
\author*[1,4]{\fnm{Deok-Sun} \sur{Lee}}\email{deoksunlee@kias.re.kr}

\affil[1]{\orgdiv{School of Computational Sciences}, \orgname{Korea Institute for Advanced Study}, \orgaddress{\street{85, Hoegi-ro},  \postcode{02455}, \state{Seoul}, \country{Korea}}}
\affil[2]{\orgdiv{Department of Physics Education}, \orgname{Seoul National University}, \orgaddress{\street{1, Gwanak-ro}, \postcode{08826}, \state{Seoul}, \country{Korea}}}
\affil[3]{\orgdiv{Center for Theoretical Physics and Artificial Intelligence Institute}, \orgname{Seoul National University}, \orgaddress{\street{1, Gwanak-ro}, \postcode{08826}, \state{Seoul}, \country{Korea}}}
\affil[4]{\orgdiv{Center for AI and Natural Sciences}, \orgname{Korea Institute for Advanced Study}, \orgaddress{\street{85, Hoegi-ro}, \postcode{02455}, \state{Seoul}, \country{Korea}}}

\abstract{
Generating graph ensembles with precisely controlled structural properties is central to investigating how network structure shapes function. Canonical ensembles impose constraints only in expectation (soft constraints), letting individual realizations fluctuate around the target, whereas enforcing hard constraints with prescribed precision in every realization remains challenging beyond fixing the degree sequence. Here we show that a reinforcement learning framework can drive a graph through degree-preserving rewirings to satisfy a prescribed assortativity, which characterizes the degree--degree correlation of adjacent nodes. By replacing the entropically dominated Metropolis--Hastings random walk with directed transport, the learned policy reduces generation cost by at least an order of magnitude while retaining over 98\% of configurational diversity. Trained on small graphs, the framework generalizes across sizes and topologies without retraining, enabling quantitative isolation of secondary observables such as the clustering coefficient. These results establish reinforcement learning as a practical paradigm for hard-constrained graph generation.
}

\maketitle

\section{Introduction}  \label{sec:intro}
The ensemble of random graphs under selected constraints provides a natural null model to assess the significance of observed features of real-world networks and to investigate how the imposed constraints influence structural and dynamical properties.
As in statistical mechanics, distinct ensembles can be employed depending on how constraints are imposed.
Canonical ensembles, formulated as exponential random graph models (ERGMs), enforce constraints only in expectation (\textit{soft constraints})~\cite{jaynes1957information,frank1986markov,park2004statistical}.
In contrast, microcanonical ensembles enforce them exactly in every realization (\textit{hard constraints}) and assign equal statistical weight to all compatible configurations~\cite{bianconi2009entropy}.
A familiar example is the degree sequence, which is constrained only on average in ERGM but fixed exactly in the configuration model~\cite{fosdick2018configuring,newman2001random}.

In finite systems, canonical ensembles introduce structural fluctuations over different realizations that can obscure the effects of the imposed constraints.
In the thermodynamic limit, inequivalence between canonical and microcanonical graph ensembles arises even in simple graphs with fixed degree sequences~\cite{squartini2015breaking,squartini2015unbiased}.
Under multiply constrained systems like the edge--triangle model~\cite{radin2013phase,chatterjee2013estimating}, canonical formulations can exhibit discontinuous phase transitions that bypass intermediate graph configurations, which remain accessible in the corresponding microcanonical ensemble.
Therefore, residual fluctuations and configurational inaccessibility under soft constraints establish a critical need for generative methods that can efficiently generate graphs under hard constraints.

In this study, we focus on the Pearson degree assortativity $\rho$, a widely used measure of degree--degree correlation of adjacent nodes~\cite{newman2002assortative,newman2003structure}.
Assortativity is a fundamental structural property, affecting percolation criticality~\cite{noh2007percolation,goltsev2008percolation}, synchronization and spreading dynamics~\cite{van2010influence,avalos2012assortative,boguna2002epidemic}, and robustness~\cite{zhou2012assortativity}.
Because Pearson assortativity can be sensitive to high-degree nodes and system size when the third moment of the degree distribution diverges, rank-based alternatives such as Spearman's rho and Kendall's tau have also been proposed~\cite{litvak2013uncovering}.
While degree-preserving rewiring, as in the configuration model, can preserve a given degree sequence, generating graph ensembles that attain a target assortativity $\rho^*$ remains challenging.
Although a sampling method that preserve a given joint-degree matrix has been proposed~\cite{bassler2015exact}, determining how to generate an ensemble of joint-degree matrices that matches a given $\rho^*$ remains open.
Meanwhile, heuristic rewirings~\cite{xulvi2004reshuffling}, like canonical ERGMs, rely on soft constraints leaving residual fluctuations and require nontrivial parameter tuning.

Given these difficulties, modern deep learning offers alternative data-driven graph generators based on
autoregressive~\cite{you2018graphrnn,liao2019efficient}, VAE~\cite{simonovsky2018graphvae,jin2018junction}, GAN~\cite{decao2018molgan,martinkus2022spectre}, and diffusion~\cite{vignac2023digress,huang2022graphgdp}.
These approaches can be steered toward desired properties via conditional generation~\cite{decao2018molgan,gomez2018automatic} or optimized for specific goals using reinforcement learning (RL), such as RNet–DQN~\cite{darvariu2021goal} and ResiNet~\cite{yang2023learning} for robust network design.
For hard-constrained graph ensemble generation, however, two obstacles dominate: representative training data that strictly satisfy constraints are scarce, and approximate conditional control is insufficient when exact compliance is required.

\begin{figure*}
\centering
\includegraphics[width=\linewidth]{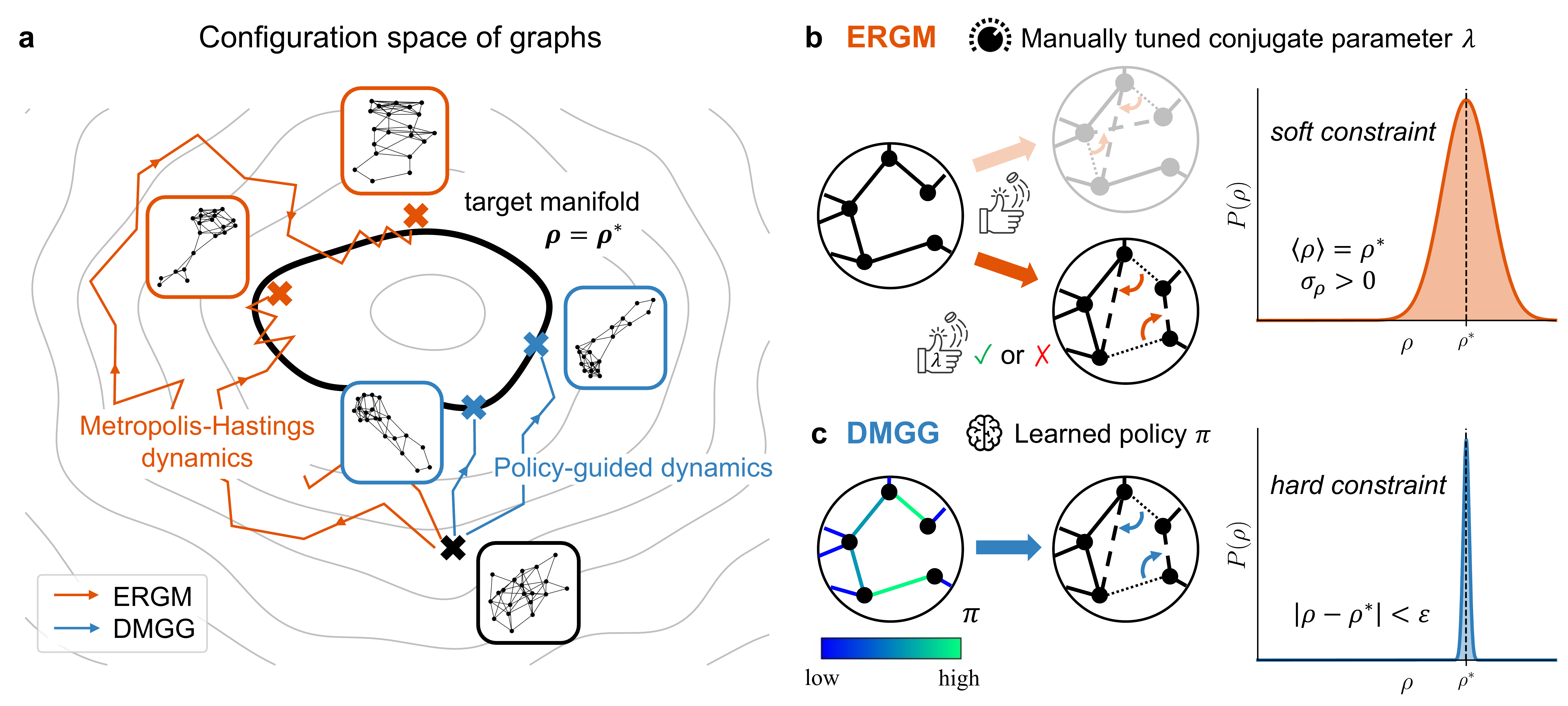}
\caption{
    \textbf{Comparison of graph generation under soft (ERGM) and hard (DMGG) assortativity $\rho$ constraints.}
    \textbf{a}, Schematic representation of ERGM and DMGG navigating the configuration space of graphs for a given degree sequence, where each contour line indicates a set of graphs sharing the same $\rho$.
    Starting from an initial graph, both ERGM and DMGG traverse the space through sequences of rewirings.
    ERGM approaches the target manifold $\{G | \rho(G)=\rho^* \}$ via stochastic Metropolis--Hastings dynamics, whereas DMGG follows policy-guided dynamics to reach the target manifold more efficiently.
    \textbf{b}, In ERGM, randomly sampled rewiring proposals are accepted or rejected according to the manually tuned conjugate parameter $\lambda$.
    The resulting graphs form a canonical ensemble satisfying the soft constraint $\langle \rho \rangle = \rho^*$, in which $\rho$ fluctuates around $\rho^*$ and yields a distribution $P(\rho)$ with finite variance $\sigma_\rho > 0$.
    \textbf{c}, In DMGG, a learned policy $\pi$ assigns scores to edges to select the rewiring that effectively navigates the graph toward $\rho^*$.
    Generated graphs satisfy the hard constraint $|\rho - \rho^*| < \varepsilon$ with user-specified tolerance $\varepsilon$, yielding a sharply localized $P(\rho)$.
} \label{fig:ergm_dmgg}
\end{figure*}

Here we introduce the \emph{Deep Macrostate Graph Generator} (DMGG), a practical framework that applies deep RL for generating graphs in a prescribed network macrostate~\cite{touchette2015equivalence,touchette2018asymptotic,fushing2014bootstrapping}, characterized by a fixed degree sequence and an assortativity $\rho$.
DMGG navigates the graph configuration space through degree-preserving rewirings and learns its policy $\pi$ solely from reward signals based on the deviation from the target $\rho$, without requiring pre-existing training graphs spanning different assortativities.
Compared with ERGM, it reduces the required rewirings by at least an order of magnitude while retaining substantial configurational diversity.
The resulting samples form a diverse hard-constrained ensemble within the target macrostate, with their distribution over compatible graph realizations induced by the learned policy.
Once trained, DMGG reliably achieves the prescribed precision across diverse assortativities, sizes, sparsities, and initial graph topologies.
We further examine the evolution of the joint-degree matrix to elucidate how DMGG achieves this accelerated generation.

\section{Results}   \label{sec:results}
\subsection{From soft to hard constraints}   \label{subsec:soft2hard}
For a simple graph $G=(\mathcal{V}, \mathcal{E})$ with $N=|\mathcal{V}|$ nodes and $E=|\mathcal{E}|$ edges, the Pearson degree assortativity considered in this work is defined as
\begin{equation}  \label{eq:assortativity}
    \rho = \frac{\frac{1}{E} \sum_{(i, j) \in \mathcal{E}} k_i k_j - \mu^2}{\frac{1}{2E} \sum_{(i, j) \in \mathcal{E}} (k_i^2 + k_j^2) - \mu^2},
\end{equation}
where $k_i$ is the degree of node $i$ and $\mu \equiv \sum_{(i, j) \in \mathcal{E}} (k_i + k_j)/2E$~\cite{newman2002assortative,newman2003mixing}.

To generate graph ensembles satisfying a soft constraint (ERGM) or a hard constraint (DMGG) on $\rho$ within the configuration space of a fixed degree sequence, we employ degree-preserving rewiring for both ERGM and DMGG.
Before applying either model, we first randomize the input graph via the configuration model to eliminate any structural biases inherited from the initial topology.
During the rewiring process, two distinct edges are chosen and their endpoints are reconnected in one of the two possible pairings, discarding moves that create self-loops, multiedges, or leave the topology unchanged.
By strictly maintaining the degree sequence, this rewire isolates the effects of degree--degree correlations from underlying degree heterogeneity.

The ERGM employs these rewirings within a Metropolis--Hastings (MH) chain~\cite{metropolis1953equation,hastings1970monte} driven by a conjugate parameter $\lambda$.
As illustrated in Fig.~\ref{fig:ergm_dmgg}\textbf{a} and \textbf{b}, randomly proposed rewiring from the current graph $G$ to a new graph $G'$ is accepted with probability
\begin{equation}
    \min[1, \exp(\lambda \Delta K)],
\end{equation}
where $K(G) \equiv \sum_{(i,j) \in \mathcal{E}} k_i k_j$ and $\Delta K \equiv K(G') - K(G)$.
Note that for a fixed degree sequence, $\rho(G)$ is linearly related to $K(G)$.
Thus, utilizing the unnormalized sum $K$ provides a direct and computationally simpler proxy for $\rho$.
This accept--reject process is repeated until the $\rho$ reaches equilibrium (see Methods).
Note that the ERGM ensemble generated from a single initial graph is microcanonical in the degree sequence but canonical in assortativity.
While theoretically well grounded, the canonical baseline is hindered by the lack of a universal mapping between $\rho$ and $\lambda$.
Consequently, applying this method requires costly parameter tuning for each degree sequence and target, and often requires an excessive number of rewirings to reach the desired state.

DMGG reformulates graph generation as a Markov decision process~\cite{bellman1957markovian,sutton1998reinforcement}, replacing the stochastic proposal and acceptance--rejection steps of MH dynamics with a policy-guided dynamics.
Starting from the randomized initial graph, a learned policy $\pi$, obtained by the training procedure described below, sequentially selects rewiring actions that efficiently navigate the graph toward the target macrostate (Fig.~\ref{fig:ergm_dmgg}\textbf{a} and \textbf{c}).
These targeted modifications continue until the graph satisfies the hard constraint $|\rho-\rho^*|<\varepsilon$.

\begin{table}
\centering
\setlength{\tabcolsep}{5pt}
\renewcommand{\arraystretch}{1.5}
\begin{tabular}{@{}lcc@{}}
    \toprule
    Property                       & Training                    & Inference      \\
    \midrule
    Size ($N$)                     & $[10^2, 10^3]$              & $[10^2, 10^4]$  \\
    Sparsity ($\langle k \rangle$) & $[3, 10]$                   & $[3, 20]$       \\
    \multirow{2}{*}{Topology}      & \multirow{2}{*}{WS, ER, BA} & WS, ER, SBM,    \\
                                    &                             & RGG, CL, HK, BA \\
    Assortativity ($\rho^*$)         & $[-0.5, 0.5]$               & $[-0.8, 0.8]$   \\
    Tolerance ($\varepsilon$)      & $0.005$                     & $0.001$         \\
    \bottomrule
\end{tabular}
\caption{
    \textbf{Training and inference domains for DMGG.}
    To reduce computational cost, training is restricted to small, sparse networks from three random graph models: Watts--Strogatz (WS), Erdős--Rényi (ER), and Barabási--Albert (BA).
    The target assortativity $\rho^*$ is sampled from a narrow range with a loose tolerance $\varepsilon$.
    The inference domain is significantly broader, encompassing larger and denser systems, a wider target range, stricter tolerances, and unseen topologies with distinct structural features, such as the stochastic block model (SBM), random geometric graph (RGG), Chung--Lu (CL), and Holme--Kim (HK) models.
}
\label{tab:domain}
\end{table}

To keep the training cost manageable, training is restricted to small, sparse Watts--Strogatz (WS)~\cite{watts1998collective}, Erdős--Rényi (ER)~\cite{erdos1960evolution}, and Barabási--Albert (BA)~\cite{barabasi1999emergence} graphs and a narrow target range with a loose tolerance (Table~\ref{tab:domain}).
These random graphs represent narrow, Poisson-like, and heavy-tailed degree distributions, respectively.
Training is performed using Proximal Policy Optimization (PPO)~\cite{schulman2017proximal} and takes approximately 4.7 hours on a single GPU for $2 \times 10^7$ training rewiring steps, corresponding to about $1.2 \times 10^5$ graph generations.
This one-time cost is amortized by reusing the same policy across graph sizes, densities, topologies, targets, and tolerances without per-instance or per-target retraining.
During inference, actions are sampled from the learned policy $\pi$ to secure diversity among generated graphs.
Notably, the policy trained with $\varepsilon=0.005$ is evaluated with tighter tolerance $\varepsilon=0.001$, demonstrating transfer to stricter user-prescribed precision without additional training.
Details regarding the RL formulation, including neural network architectures, rewards, and training procedures, are provided in Methods.
We also benchmark DMGG against a target-directed greedy heuristic and a target-conditioned adaption of ResiNet~\cite{yang2023learning}.
Their diversity, generation efficiency, and computational costs are compared in Appendix~\ref{app:baseline}.

\begin{figure}
    \centering
    \includegraphics[width=\linewidth]{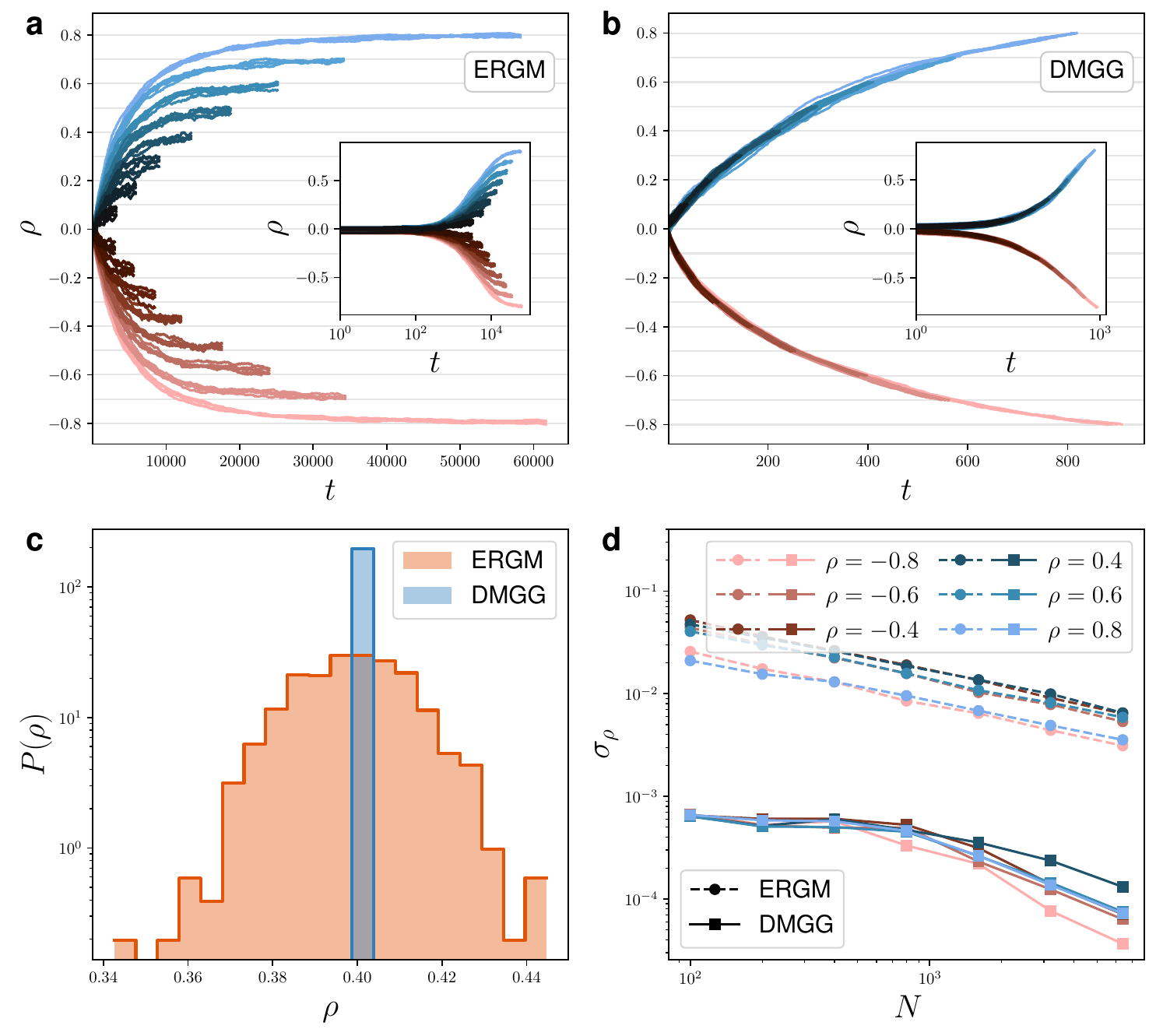}
    \caption{
        \textbf{Convergence and precision in assortativity-constrained graph generation.}
        \textbf{a},\textbf{b}, Evolution of assortativity $\rho$ during the generation process of ERGM and DMGG, respectively.
        Starting from an initial Erdős--Rényi (ER) graph ($N=1000$ and $E=3000$) with neutral assortativity ($\rho \approx 0$), five trajectories for each target (indicated by color) are shown.
        Insets illustrate the same trajectories on a logarithmic time scale.
        \textbf{c}, Distribution of $\rho$ measured over $1000$ generated graphs with target assortativity $\rho^* = 0.4$.
        \textbf{d}, Standard deviation of assortativity $\sigma_\rho$ versus system size $N$, with the same color scheme as in \textbf{a} and \textbf{b}.
        Dashed lines with circle markers and solid lines with square markers correspond to ERGM and DMGG, respectively.
        Results are shown for ER graphs in which the mean degree is fixed at $6$ across all system sizes.
    } \label{fig:trajectory}
\end{figure}

We first compare DMGG with ERGM when the initial graph is an ER graph with $N=1000$ and $E=3000$.
Both methods eventually approach the prescribed assortativities in the range of $[-0.8, 0.8]$, but their paths differ qualitatively.
Fig.~\ref{fig:trajectory}\textbf{a} and \textbf{b} compare representative trajectories of $\rho$ as a function of the number of rewirings $t$ during the generation process.
Because the ERGM follows MH dynamics governed by a conjugate parameter $\lambda$ specific to each target, it exhibits distinct trajectories.
In contrast, DMGG uses a single policy $\pi$ whose dominant control signal is the direction toward the target, giving rise to similar macroscopic trajectories of $\rho$ for targets with the same sign while retaining stochasticity in the underlying rewiring sequences.
This offers a significant practical advantage, as a single generation process can simultaneously generate graphs with a range of $\rho$.

Fig.~\ref{fig:trajectory}\textbf{c} compares the distribution $P(\rho)$ of the generated graphs for the $\rho^* = 0.4$ case.
The ERGM satisfies only soft constraints, producing a broad distribution centered near the target.
In contrast, the distribution from DMGG is sharply concentrated around the target value with tolerance $\varepsilon = 0.001$.
As shown in Fig.~\ref{fig:trajectory}\textbf{d}, the standard deviation $\sigma_\rho$ of the canonical ensemble decreases with system size $N$ at a fixed mean degree $\langle k \rangle = 6$~\cite{touchette2015equivalence}.
For DMGG, $\sigma_\rho$ is inherently bounded by the tolerance $\varepsilon$ and decreases further as $N$ increases, remaining smaller than that of ERGM.
This is because $\pi$ is trained to minimize the gap $|\rho - \rho^*|$ even within the tolerance window, and larger systems allow for finer control as each rewiring modifies $\rho$ by only $O(E^{-1})$.

\subsection{Ensemble diversity with accelerated generation} \label{subsec:diversity_efficiency}
An ensemble generator would be of limited scientific value if it collapsed onto only a few specific realizations rather than exploring a substantial portion of the accessible configuration space.
We therefore examine whether DMGG retains sufficient configurational diversity while remaining practically efficient.

\begin{figure}
\centering
\includegraphics[width=\linewidth]{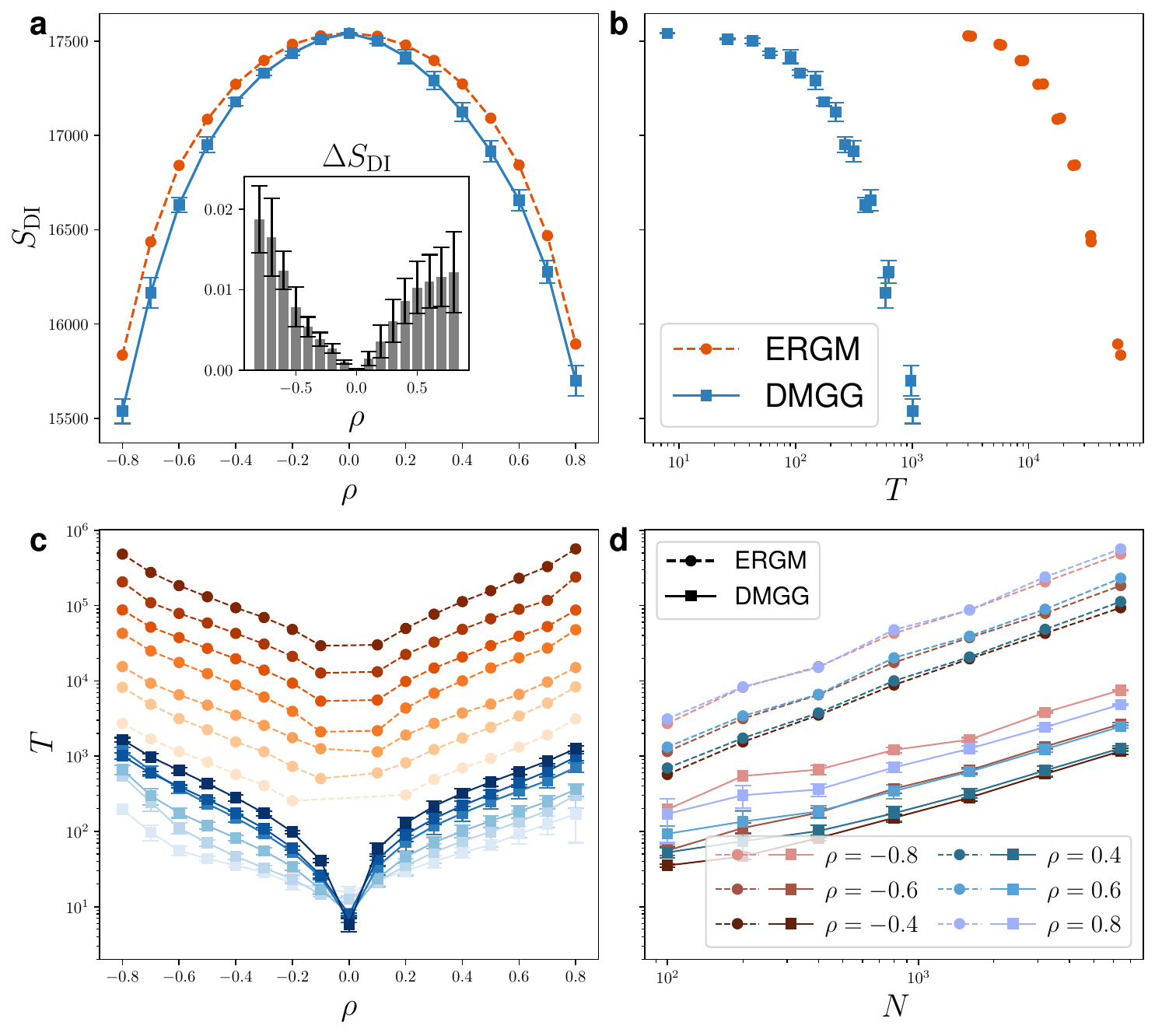}
\caption{
    \textbf{Configurational diversity and computational efficiency of DMGG.}
    \textbf{a}, Dyad-independent entropy $S_\text{DI}(\{p_{ij}\})$ measured in the ensembles generated by ERGM and DMGG, as a function of assortativity $\rho$.
    The inset shows the deviation for each $\rho$, $\Delta S_\text{DI} \equiv \left( S^{\text{ERGM}}_\text{DI}-S^{\text{DMGG}}_\text{DI} \right) / S^{\text{ERGM}}_\text{DI}$.
    \textbf{b}, Parametric plot of $S_\text{DI}$ as a function of the number of required rewirings $T$, where each marker corresponds to the data for a distinct assortativity in \textbf{a}.
    \textbf{c}, $T$ as a function of assortativity $\rho$ with system sizes $N=100, 200, \dots, 6400$ indicated by color gradient (light to dark).
    \textbf{d}, Scaling of $T$ with system size $N$ for fixed targets (colors).
    Dashed lines with circle markers and solid lines with square markers correspond to ERGM and DMGG, respectively.
    DMGG results are averaged over five independently trained policies, with error bars indicating 95\% confidence intervals across the training runs.
    All results are shown for ER graphs with a fixed mean degree $\langle k \rangle = 6$.
} \label{fig:entropy_efficiency}
\end{figure}

Since exhaustive graph construction and direct estimation of Shannon entropy is intractable for large graph configuration spaces, we quantify ensemble diversity using the dyad-independent entropy:
\begin{equation} \label{eq:entropy}
    S_\text{DI} \equiv -\sum_{i < j} \left[ p_{ij} \ln p_{ij} + (1-p_{ij}) \ln (1-p_{ij}) \right],
\end{equation}
where $p_{ij}$ denotes the marginal probability that an edge exists between nodes $i$ and $j$.
By assuming statistical independence between edges, $S_\text{DI}$ provides an upper bound on the Shannon entropy of degree-constrained ensembles, as the fixed degree sequence can impose correlations.
This proxy quantifies whether the generated graphs collapse onto a narrow subset of configurations but does not establish uniform sampling over the constrained graph space.
The relation between $S_\text{DI}$ and the exact Shannon entropy is examined by exhaustive enumeration of a small graph in Appendix~\ref{app:small}.

Fig.~\ref{fig:entropy_efficiency}\textbf{a} shows the $S_\text{DI}$ from ERGM and DMGG ensembles starting from a single initial ER graph ($N=1000$ and $E=3000$).
We evaluated five independently trained DMGG and report the mean with 95\% confidence intervals across training runs.
For each trained policy and target assortativity, we generate 1000 graph instances from separate runs and measure $p_{ij}$.
The estimated $S_\text{DI}$ decreases as $|\rho|$ increases, consistent with the expectation that the accessible graph space shrinks under stronger constraints.
As DMGG enforces a hard constraint, its diversity is lower than that of the ERGM.
Crucially, however, the DMGG and ERGM curves nearly overlap, with a deviation $\Delta S_\text{DI} \equiv \left( S^{\text{ERGM}}_\text{DI}-S^{\text{DMGG}}_\text{DI} \right) / S^{\text{ERGM}}_\text{DI}$ remaining below $2\%$ across the evaluated range $\rho\in[-0.8,0.8]$ (inset).

\begin{figure*}
    \centering
    \includegraphics[width=\linewidth]{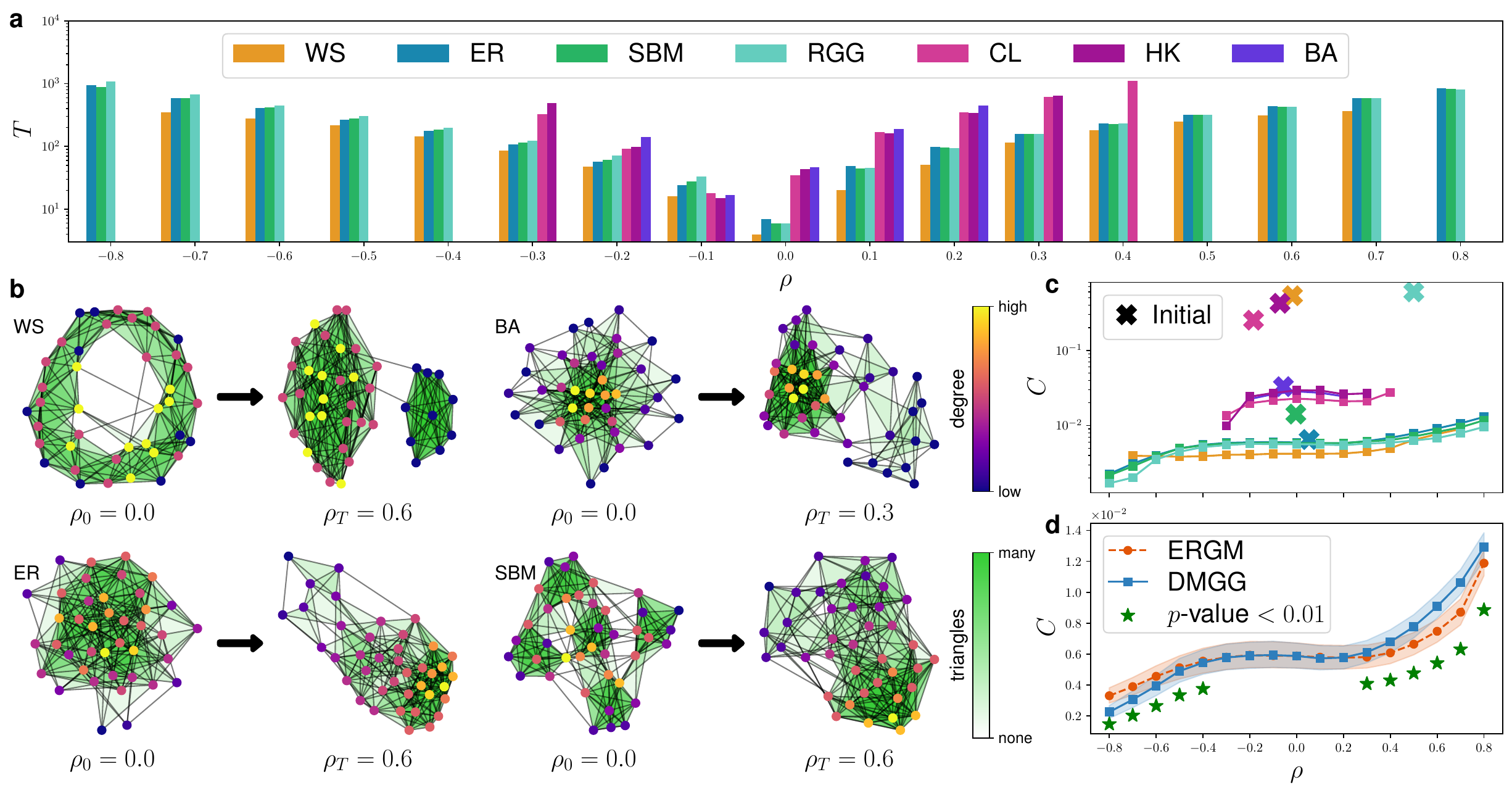}
    \caption{
        \textbf{Generalization of DMGG across diverse graph families.}
        \textbf{a}, Number of rewirings $T$ required to reach assortativity $\rho$ in different initial topologies, all with $N=1000$ and $E \approx 3000$.
        Topology labels and colors denote the graph families used to generate the initial degree sequences.
        Missing bars indicate target $\rho$ values outside the empirically estimated feasible range of a given topology.
        \textbf{b}, Representative examples of graph transformation by DMGG.
        The initial graphs (left) with diverse topologies with assortativity $\rho_0 \approx 0$ are rewired into target configurations (right) with prescribed assortativities $\rho_T$.
        Node colors indicate relative degrees, and triangles contributing to the clustering coefficient $C$ are shaded green.
        \textbf{c}, Average $C$ of the generated ensembles from DMGG as a function of $\rho$, using the same color scheme as in \textbf{a} to denote topology.
        $(\rho, C)$ of the initial graphs are marked by crosses.
        \textbf{d}, $C(\rho)$ for ERGM and DMGG ensembles generated from the same initial ER graph ($N=1000$ and $E=3000$).
        Shaded areas indicate the interquartile range, and green stars mark $\rho$ where the difference in $C$ between two ensembles is statistically significant ($p<0.01$).
    } \label{fig:generalization}
\end{figure*}

Fig.~\ref{fig:entropy_efficiency}\textbf{b} demonstrates that DMGG reaches this high-diversity regime with significantly fewer rewirings than the ERGM, quantified by the required number of rewirings $T$ (see Methods).
Thus, in DMGG, hard constraint satisfaction, configurational diversity, and efficient generation are achieved simultaneously.
Together with the baseline comparisons, DMGG provides a balanced combination of rewiring efficiency, configurational diversity, and scalable action selection, rather than minimizing $T$ alone.
Robustness across independently sampled initial graphs and the roles of key policy components are examined in Appendix~\ref{app:robustness}.

We systematically quantify this required number of rewiring steps using ER graphs with a fixed mean degree $\langle k \rangle = 6$.
As in the preceding analyses, the results are aggregated over five independently trained policies.
As shown in Figs.~\ref{fig:entropy_efficiency}\textbf{c} and \textbf{d}, DMGG consistently requires at least an order of magnitude fewer rewirings $T$ than the ERGM across $\rho \in [-0.8, 0.8]$ and system sizes $N=100, 200, \dots, 6400$.
We fit the generation cost $T(\rho, N)$ as
\begin{equation} \label{eq:scaling}
    T \sim 10 ^{\alpha |\rho|} N^\beta,
\end{equation}
with exponents $\alpha_\text{ERGM} = 1.51 \pm 0.08$ and $\beta_\text{ERGM} = 1.14 \pm 0.29$ for the ERGM, compared with $\alpha_\text{DMGG} = 1.63 \pm 0.30$ and $\beta_\text{DMGG} = 0.93 \pm 0.05$ for DMGG.
The exponential dependence on $|\rho|$ reflects the extensive structural reorganization required to reach extreme assortativities.
Meanwhile, the near-linear scaling with $N$ is theoretically expected, as a single rewiring adjusts $\rho$ by $O(E^{-1})$.

\subsection{Generalization across topologies}   \label{subsec:generalization}

Beyond the scale-up generalization, we examine transfer across initial graph topologies using graph families spanning a broad range of degree heterogeneity.
Specifically, we consider (1) \textit{narrow distributions} represented by WS small-world graphs; (2) \textit{Poisson-like distributions}, including ER random graphs, stochastic block models (SBM)~\cite{holland1983stochastic} with community structure, and spatially clustered random geometric graphs (RGG)~\cite{dall2002random}; and (3) \textit{heavy-tailed distributions}, encompassing scale-free BA graphs, Chung--Lu (CL) models~\cite{chung2002average} with controllable degree exponent, Holme--Kim (HK) models~\cite{holme2002growing} with high clustering.
Throughout, target assortativities are chosen within the empirically estimated feasible range of each degree sequence.
Unlike ERGMs, which require exhaustive, case-by-case parameter tuning for each initial topology and system size, a single pre-trained DMGG model can be applied across all these diverse structures without any retraining.

As initial graphs are randomized via the configuration model and their assortativity starts near zero, the number of rewiring steps $T$ required to reach an assortativity naturally increases with $|\rho|$ (Fig.~\ref{fig:generalization}\textbf{a}).
Notably, the $T(\rho)$ profiles are primarily governed by the degree distribution.
While the WS model with the narrowest degree distribution requires the fewest rewirings, models with Poisson-like distributions (ER, SBM, and RGG) exhibit slightly higher $T$.
Graphs characterized by broad, heavy-tailed degree distributions (CL, HK, and BA) demand significantly more rewirings.

\begin{figure*}
    \centering
    \includegraphics[width=\linewidth]{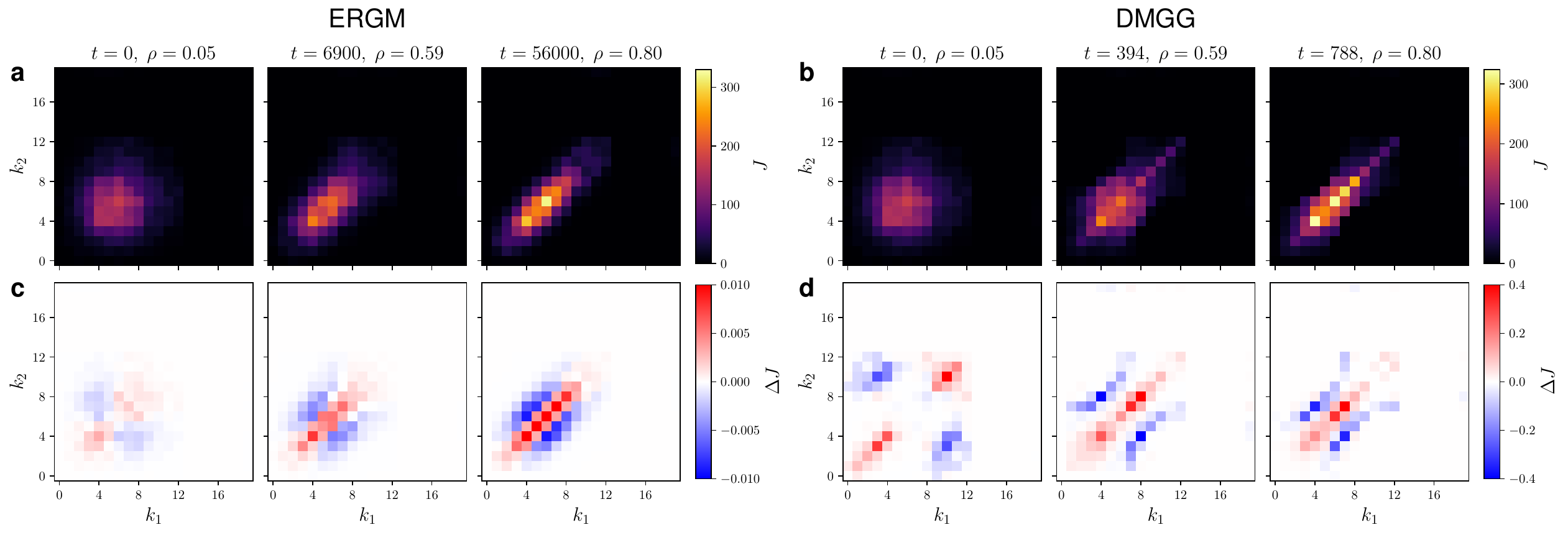}
    \caption{
        \textbf{Rewiring strategies of ERGM and DMGG.}
        Snapshots of the joint-degree matrix $J$ (top row) and its expected change per rewiring $\Delta J$ (bottom row) during the generation process with target $\rho^*=0.8$.
        \textbf{a}, \textbf{c}, ERGM dynamics driven by the Metropolis--Hastings process.
        \textbf{b}, \textbf{d}, DMGG dynamics driven by a policy-guided process.
    }   \label{fig:mixing}
\end{figure*}

To visually demonstrate structural evolution, Fig.~\ref{fig:generalization}\textbf{b} shows representative transformations of small graphs ($N=40$ and $E = 200$).
As DMGG drives the networks toward high assortativity, high-degree nodes converge into tightly interconnected central cores.
Furthermore, networks sharing similar degree distributions, such as ER and SBM, evolve into comparable structures.

By completely eliminating the computational bottleneck of parameter tuning, this generalizability establishes DMGG as a highly versatile framework for generating hard-constrained null models.
Consequently, it enables systematic investigations into how primary constraints, such as the degree sequence and $\rho$, influence secondary structural observables.
As a demonstration, Fig.~\ref{fig:generalization}\textbf{c} compares the average clustering coefficient $C$ of various initial graphs and their corresponding DMGG-generated ensembles as a function of $\rho$.
Within the generated ensembles, the $C(\rho)$ curves group distinctly according to the degree distribution.
Specifically, ensembles constructed from heavy-tailed degree sequences (CL, HK, and BA) exhibit substantially higher $C$ than those with narrower degree distributions.
Within each group, $C$ also tends to be higher at more positive $\rho$.
Furthermore, while the ensembles provide a practical baseline for $C$ under the prescribed degree sequence and assortativity, the initial graphs often display higher $C$ driven by their specific generative mechanisms, such as spatial embeddings (RGG) or explicit triad formation (HK).
DMGG thus enables quantitative assessment of structural features that are not reproduced by the ensemble constrained by the prescribed macrostate.

On the other hand, Fig.~\ref{fig:generalization}\textbf{d} highlights the impact of constraint stringency by comparing ERGM and DMGG ensembles initialized from a single ER graph.
While both ensembles share similar $C(\rho)$ trends, we observe that the variation of $C$ is more pronounced under DMGG.
The differences between the two ensembles are statistically significant across the large $|\rho|$ regime ($p<0.01$), indicating that discrepancies between soft- and hard-constrained ensembles become robust when strong constraints are enforced.

The observed generalization across graph sizes and initial topologies can be understood through two complementary properties of the DMGG.
Within a fixed degree sequence, the direction and relative effect of a valid rewiring on $\rho$ are determined by the degrees of its four endpoints.
By applying shared parameters across nodes, the GNN can learn and exploit this local relation when evaluating candidate rewirings in graphs of different sizes and topologies.
As $N$ increases, the same local decision rule remains applicable, increasing the number of required updates, consistent with the $O(E^{-1})$ change per rewiring and the near-linear $T(N)$ scaling in Eq.~\eqref{eq:scaling}.
Likewise, when the initial topology changes, the set of valid rewirings available at each step changes, but the policy can evaluate them using the same learned rule without topology-specific retraining.

\subsection{Directed flux in the configuration space}   \label{subsec:flux}
To trace the origin of DMGG's speedup, we investigate the joint-degree matrix $J$, where $J_{k_1k_2}$ denotes the number of edges connecting nodes of degrees $k_1$ and $k_2$~\cite{bassler2015exact}.
We define the flux $\Delta J$ as the expected change in $J$ per rewiring step: $\Delta J_{k_1k_2}(t) \equiv \langle J_{k_1k_2}(t+1) - J_{k_1k_2}(t) \rangle$.
Snapshots of $J$ and $\Delta J$ during the ERGM and DMGG generation processes for a target $\rho^*=0.8$ are shown in Fig.~\ref{fig:mixing}.

As illustrated in Figs.~\ref{fig:mixing}\textbf{a} and \textbf{b}, both ERGM and DMGG traverse similar sequences of $J$ matrices, confirming that they reach comparable macrostates.
However, the fluxes $\Delta J$ shown in Figs.~\ref{fig:mixing}\textbf{c} and \textbf{d} reveal fundamentally distinct strategies.
Qualitatively, ERGM proposals are drawn uniformly from all admissible rewires.
This inherently concentrates $\Delta J$ around the bright region in Fig.~\ref{fig:mixing}\textbf{c}, where the joint-degree distribution is peaked, representing the entropic bias.
In contrast, DMGG bypasses this by explicitly targeting high-impact, off-diagonal degree pairs that produce large changes in $\rho$.
Quantitatively, the ERGM flux is attenuated by stochastic fluctuations and MH rejections, whereas the learned policy in DMGG induces more directed flux, approximately 40 times larger in magnitude of $\Delta J$, thereby reaching the target assortativity in an order of magnitude fewer rewirings.

This mechanism clarifies that the acceleration arises from a fundamental shift in the navigation dynamics within the configuration space.
The ERGM approaches the target via an entropically dominated random walk, spending significant time exploring abundant but low-impact local moves.
In contrast, DMGG executes a directed transport, effectively identifying the sparse subset of rewirings that constitute the most efficient route to the target manifold.
Consequently, the model achieves efficiency by replacing a slow diffusive relaxation with a directed transport driven by the learned policy.

\section{Discussion}    \label{sec:discussion}
The central advance of DMGG is to establish a practical route to hard-constrained graph ensemble generation, where a single policy produces diverse graphs compatible with a prescribed macrostate across a wide range of graph sizes and topologies.
Under the diagnostics used here, the resulting ensembles retain substantial configurational diversity, and Fig.~\ref{fig:generalization}\textbf{d} provides a proof of principle that soft- and hard-constrained ensembles can exhibit visibly different trends in a secondary observable such as clustering coefficients.

Although we focus on an assortativity constraint with degree-preserving rewiring, the same framework could in principle be adapted to other graph properties, provided that the target observable can be evaluated efficiently and that a valid action set is available to modify it.
This includes rank-based assortativity measures, local observables such as the average clustering coefficient and motif counts, or global metrics such as diameter and algebraic connectivity.
Extensions to multiple constraints would offer a route to probe regimes where ERGM struggles with phase coexistence and ensemble non-equivalence.
Furthermore, applications to enriched graph classes, such as bipartite, spatially embedded, multilayer graphs, and hypergraphs, remain promising but would require corresponding constraint evaluations and action designs.

\section{Methods}   \label{sec:method}

\subsection{Canonical ensemble via ERGM}    \label{subsec:canonical}
ERGM defines a graph distribution $P(G)$ on a graph space $\G$ by maximizing the Shannon entropy subject to soft constraints $\langle C_i(G)\rangle=c_i$, yielding
\begin{equation} \label{eq:ergm}
    P(G)=\frac{1}{Z}\exp\left(\sum_i \lambda_i C_i(G)\right),
\end{equation}
where $\lambda_i$ are conjugate parameters for the corresponding constraints $C_i$ and $Z$ is the partition function.

To sample from Eq.~\eqref{eq:ergm} under the assortativity constraint, we employ the Metropolis--Hastings (MH) algorithm.
In each step, a degree-preserving rewiring is proposed by selecting two edges, $(u, v)$ and $(x, y)$ uniformly at random and rewiring them to either $\{(u, x), (v, y)\}$ or $\{(u, y), (v, x)\}$, denoted by modes $b \in \{0, 1 \}$.
As described in the main text, this proposal is accepted with probability $\min[1, \exp(\lambda \Delta K)]$.
By evaluating the unnormalized increment $\Delta K$ rather than the exact $\Delta \rho$, the MH algorithm targets the assortativity constraint while avoiding the computation of system-size dependent normalization factors.

The number of required rewirings $T$ in ERGM is defined as the duration of the transient regime.
We estimate it by identifying the time when the smoothed ensemble average of $\rho$ settles within the fluctuation band of the steady state.
In this sense, $T$ is the time when the initial transient bias has decayed below the level of stationary fluctuations.

For every ERGM result plotted as a function of $\rho$ in the main text, we first tune the conjugate parameter $\lambda$ such that $\langle \rho \rangle$ matches the target assortativity.
Reported statistics are derived from equilibrium configurations at the corresponding $\lambda$.
Accordingly, for ERGM, $\rho$ should be understood as the expected value, unlike the realized value in DMGG, which is constrained to lie within the target tolerance window.

To estimate the feasible assortativity interval associated with a given degree sequence, we use an empirical search.
Starting from a given graph of $E$ edges, we repeatedly propose random rewirings and accept only those that increase $\rho$; after $50E$ rewiring attempts, the resulting assortativity is taken as an empirical estimate of the maximum feasible value.
Repeating the same procedure while accepting only moves that decrease $\rho$ gives the corresponding empirical minimum.
This procedure yields a conservative estimate of the feasible interval, and it may exclude attainable near-extremal values.
A more extensive search could expand the estimated interval but would not affect the results at the target values already evaluated.

\subsection{DMGG architecture and training} \label{subsec:dmgg}
DMGG casts the generation of graphs compatible with a prescribed macrostate as a sequential rewiring process guided by a learned policy.
At step $t$, the graph $G_t$ is represented by its edge list and normalized node degrees $k_i/k_{\max}$.
This representation proves sufficient for the task, avoiding the computational overhead associated with complex graph encodings.
Crucially, the policy network is conditioned solely on the sign of the gap, $\text{sgn}[\rho^*-\rho(G_t)]$, a design choice that enables the model to extrapolate to target assortativities beyond the training region.
In contrast, the value network receives the full gap to estimate the effective distance to the target, quantifying the remaining rewirings to reach the target manifold.

A rewiring action $a_t = (e_1, e_2, b)$ involves selecting two distinct edges $e_1, e_2$ and a rewiring mode $b$.
Since the number of valid rewirings scales as $O(E^2)$, the resulting combinatorial explosion renders direct scoring of all possible rewirings computationally intractable.
We therefore adopt a factorized policy architecture~\cite{tavakoli2018action}:
\begin{equation} \label{eq:factorized}
    \pi(a_t | G_t) = \pi_1(e_1 | G_t) \cdot \pi_2(e_2 | e_1, G_t) \cdot \pi_b(b | e_1, e_2, G_t).
\end{equation}
Each conditional distribution is parametrized by a separate neural network, reducing the output dimension from quadratic to linear in the number of edges.

To ensure physical validity, we implement action masking~\cite{huang2020closer}, setting the probabilities of invalid rewirings to zero before sampling the action.
Furthermore, rewirings that leave $K(G)$ unchanged are also masked, as they do not contribute to controlling $\rho$.
If this restriction empties the action set, the mask is relaxed to enforce only validity constraints.

A graph neural network built from GIN layers~\cite{xu2018how} with FiLM conditioning~\cite{perez2018film} extracts hidden features from $G_t$ and $\rho^*$, which are then fed to the three policy heads and value network.
Target assortativities are sampled from the feasible interval of the given degree sequence with a small margin from the extremal values.
We implement a physically confined tolerance threshold
\begin{equation}   \label{eq:epsilon_phys}
    \varepsilon_{\rm phys} \equiv \max \left(\varepsilon, \frac{1}{E \cdot \text{Var}(k)}\right),
\end{equation}
where the second term accounts for the minimum change of $\rho$ by a single rewiring in finite systems.
This correction dominates only in small graphs where single rewiring steps induce changes comparable to the given $\varepsilon$.

The learning process is driven by a reward signal that the policy seeks to maximize.
However, relying solely on a binary indicator for hard constraint satisfaction introduces the fundamental challenge of reward sparsity~\cite{pathak2017curiosity}.
To address this, we utilize reward shaping based on a potential function defined in the configuration space~\cite{ng1999policy}:
\begin{equation}    \label{eq:potential}
    \phi_\zeta(\rho, \rho^*) \equiv \frac{|\rho^*| + \zeta}{|\rho^*-\rho| + \zeta},
\end{equation}
where $\zeta > 0$ controls the gradient sharpness near the target $\rho^*$.
This potential is normalized such that $\phi_\zeta(0, \rho^*) \approx 1$ for initial graphs.
For notational simplicity, we denote $\phi_\zeta(\rho, \rho^*)$ as $\phi(\rho)$.
The reward $r_t$ at step $t$ is then defined as
\begin{equation} \label{eq:reward}
    r_t = \left[ \phi(\rho_t) - \phi(\rho_{t-1}) \right] - p + s \cdot \mathds{1}_{[\rho^* - \varepsilon, \rho^* + \varepsilon]}(\rho_t),
\end{equation}
where $p>0$ is a step penalty, $s \gg 1$ is a success bonus, and $\mathds{1}_A(\cdot)$ represents the indicator function.
The potential difference term provides dense feedback to guide the policy toward the target, while the step penalty encourages rapid convergence.
Ultimately, the policy is optimized to maximize the cumulative reward $\sum_{t=0}^T \gamma^t r_t$, where the discount factor $\gamma \in (0,1)$ balances immediate and future rewards.
Training was performed using PPO with $\gamma = 0.997$, $\zeta = 0.005$, $p = 0.001$, and $s = 100$.

All experiments were conducted on an AMD Ryzen Threadripper 9970X CPU and an NVIDIA RTX PRO 6000 Blackwell Workstation Edition GPU.

\appendix
\setcounter{figure}{0}
\setcounter{table}{0}
\renewcommand{\thefigure}{A\arabic{figure}}
\renewcommand{\thetable}{A\arabic{table}}
\renewcommand{\theHfigure}{appendix.\arabic{figure}}
\renewcommand{\theHtable}{appendix.\arabic{table}}

\section{Sensitivity analysis of DMGG}  \label{app:robustness}

\begin{figure}
    \centering
    \includegraphics[width=\linewidth]{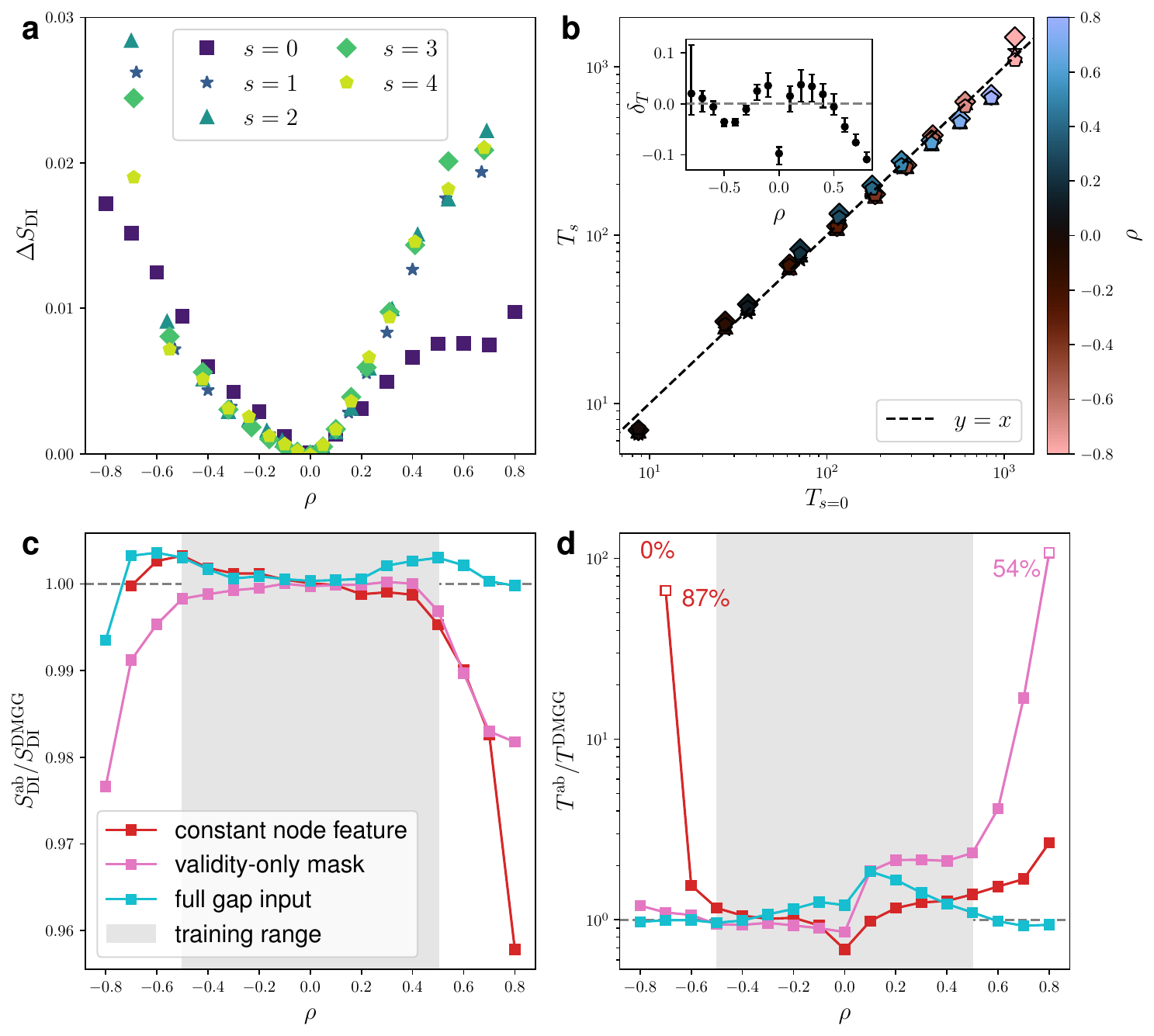}
    \caption{
        \textbf{Sensitivity of DMGG to initial graphs and policy ablations.}
        All results use ER graphs with $N=1000$ and $E=3000$.
        \textbf{a}, Relative dyad-independent entropy deviation $\Delta S_\text{DI}$ from matched ERGM ensembles for five initial graphs $s=0,1,\dots,4$.
        \textbf{b}, Mean rewiring count $T_s$ for $s=1,\dots,4$ versus that for the reference graph, $T_{s=0}$.
        Marker shapes identify $s$ as in \textbf{a}, and colors represent $\rho$.
        The inset shows $\delta_T \equiv \log_{10}(T_s/T_{s=0})$, where points indicate medians and whiskers span the min--max range over $s=1,\ldots,4$.
        \textbf{c}, \textbf{d}, Ratios of dyad-independent entropy and rewiring count, respectively, for each ablated policy to DMGG.
        Dashed lines indicate equality with the original DMGG, and shading marks the training range.
        Open markers and adjacent percentages in \textbf{d} indicate success rates below $100\%$.
        When no run succeeds, only the percentage is shown without the marker.
    } \label{fig:sensitivity}
\end{figure}

We complement the main results with two sensitivity analyses.
Fig.~\ref{fig:sensitivity}\textbf{a} and \textbf{b} examine whether a single trained policy transfers across initial graphs, while Fig.~\ref{fig:sensitivity}\textbf{c} and \textbf{d} isolate the roles of three policy design choices.

\subsection{Initial graphs}   \label{app:robust_initial}

We apply a single trained DMGG policy to five independently generated ER graphs, denoted by $s=0, 1, \dots, 4$, with $s=0$ serving as the reference.
Each graph has a distinct degree sequence, and at least 1000 samples are generated for each $\rho$.
Fig.~\ref{fig:sensitivity}\textbf{a} shows that the relative dyad-independent entropy deviation $\Delta S_\text{DI}$ remains below $3\%$ for every initial graph.
In Fig.~\ref{fig:sensitivity}\textbf{b}, the number of rewiring steps closely follow the identity line $T_s=T_{s=0}$, with a log-scale Pearson correlation of $r=0.992$.
The inset further shows that $\delta_T \equiv \log_{10}(T_s/T_{s=0})$ fluctuates around zero with only modest variation across initial graphs.
Thus, both the ensemble diversity and rewiring efficiency reported in Fig.~\ref{fig:entropy_efficiency} persist across independently sampled initial graphs.

\subsection{Policy ablations} \label{app:ablation}

For the ablation study, we consider three controlled modifications of DMGG, each trained from scratch using the procedure in Sec.~\ref{subsec:dmgg}.
(i) \emph{Constant node feature}: we replace the normalized degree feature of every node with a constant value of one, while retaining GNN message passing and the original action masks.
(ii) \emph{Validity-only mask}: we mask only those actions that would create self-loops or multiedges, allowing all valid actions that leave $K(G)$ unchanged.
(iii) \emph{Full gap input}: the policy receives the full gap $\rho^*-\rho(G_t)$ rather than only its sign.

Fig.~\ref{fig:sensitivity}\textbf{c} and \textbf{d} report ensemble diversity and rewiring steps relative to the original DMGG, so unity indicates unchanged performance.
An episode is successful if it reaches the target within $10^5$ rewiring steps, and the reported $T$ is averaged over successful episodes.
Under the \emph{constant node feature}, no episode succeeds at $\rho=-0.8$, and reaches $87\%$ success rate at $\rho=-0.7$.
Its rewiring cost increases toward both boundaries, while its largest diversity loss occurs at $\rho=0.8$.
The \emph{validity-only mask} exhibits similar boundary sensitivity, with reduced diversity and increased rewiring costs at both extremes and a $54\%$ success rate at $\rho=0.8$.
The \emph{full gap input} supplies the magnitude as well as the direction of the target gap.
This finer conditioning is consistent with slightly higher diversity within the training interval $\rho^* \in [-0.5,0.5]$, although the gain diminishes beyond this range and is generally accompanied by a modest increase in rewiring cost within it.

Overall, the normalized degree node feature and masking of ineffective moves primarily improve boundary robustness, whereas the gap conditioning exhibit modest, target-dependent tradeoffs.

\section{Exact entropy analysis in a small system}  \label{app:small}

For sufficiently small graphs, we can exhaustively enumerate all graphs with a fixed degree sequence.
Let $\Omega(K^*)$ be the number of graphs satisfying $K(G)=K^*$.
The microcanonical Shannon entropy is then $S^\text{micro}(K)=\ln \Omega(K)$, while the exact edge marginals $p_{ij}^\text{micro}(K)$ determine $S_\text{DI}^\text{micro}$ through Eq.~\eqref{eq:entropy}.
The enumeration also yields the exact canonical statistics.
For a target $K^*$, the conjugate parameter $\lambda$ is determined by solving $\langle K \rangle = \sum_K K P_\lambda(K)=K^*$, where $P_\lambda(K)=\Omega(K) \exp(\lambda K)/Z(\lambda)$.
The canonical Shannon entropy is then $S^\text{canonical}=\ln Z(\lambda)-\lambda K^*$.
Finally, the edge marginals are given by $p_{ij}^\text{canonical}=\sum_K P_\lambda(K)p_{ij}^\text{micro}(K)$, from which $S_\text{DI}^\text{canonical}$ is calculated.

\begin{figure}
    \centering
    \includegraphics[width=\linewidth]{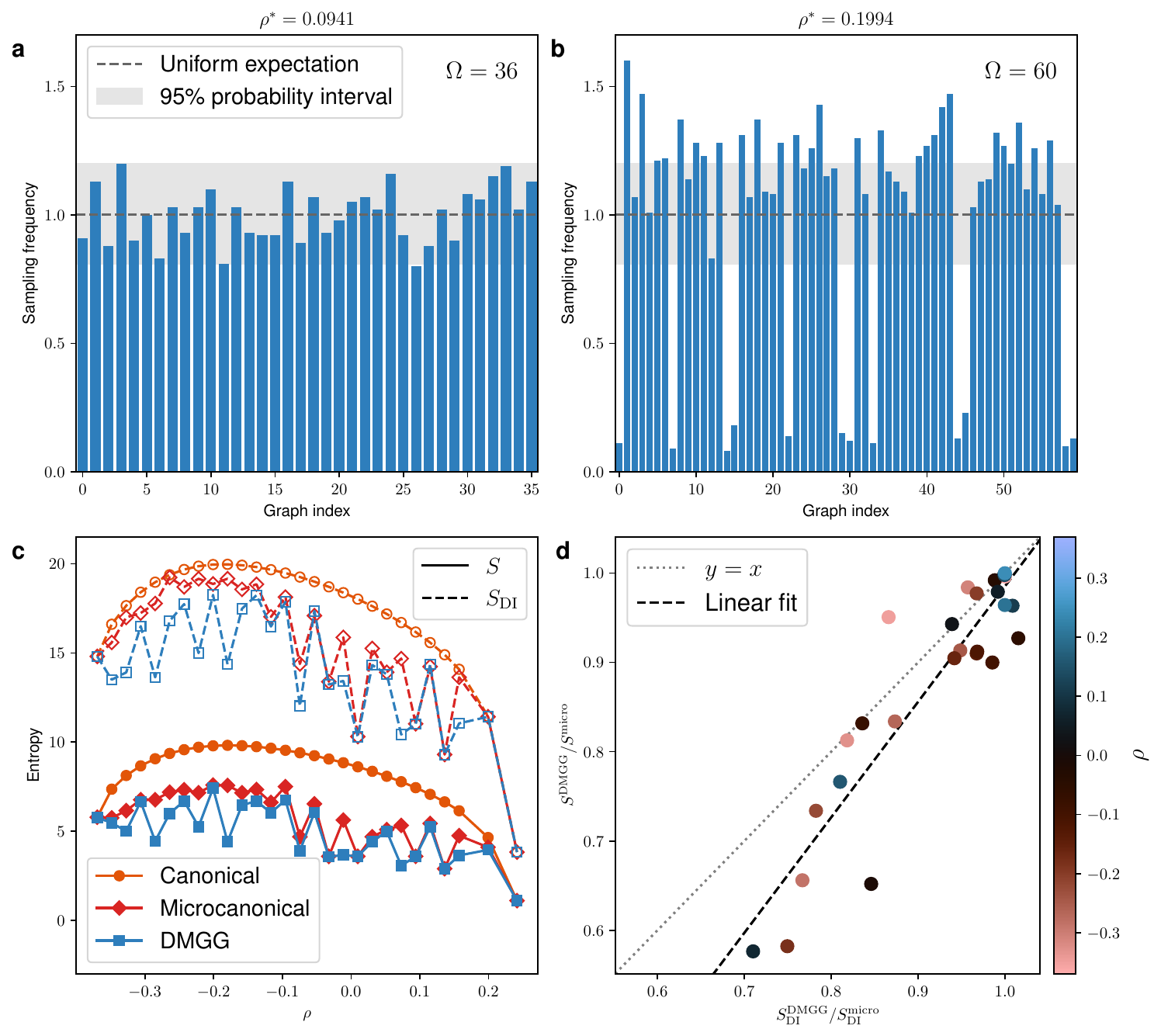}
    \caption{
        \textbf{Sampling distributions and entropy analysis in a small system.}
        \textbf{a,b}, DMGG sampling distributions at $\rho^*=0.0941$ and $0.1994$, respectively.
        Counts for the $\Omega$ feasible graphs, indexed in an arbitrary order, are normalized by their nominal uniform expectation.
        Dashed lines mark unity and shaded bands show pointwise 95\% probability intervals under independent uniform sampling.
        \textbf{c}, Exact canonical and microcanonical entropies and sample-based DMGG estimates.
        Solid lines with filled markers represent $S$, and dashed lines with open markers represent $S_\text{DI}$.
        \textbf{d}, Entropy ratio between DMGG and microcanonical ensemble for $S$ and $S_\text{DI}$, colored by $\rho$.
        Dotted and dashed lines show $y=x$ and a linear fit, respectively.
    }   \label{fig:small}
\end{figure}
We consider an ER graph with $N=10$ and mean degree $\langle k \rangle=6$, whose degree sequence is $(7,7,6,6,7,7,6,8,4,2)$.
There are 18,260 graphs with this degree sequence, distributed over 28 feasible assortativity values ranging from $-0.369$ to $0.242$.
For each feasible $K$, we generate $100\,\Omega(K)$ DMGG samples.
Fig.~\ref{fig:small}\textbf{a} and \textbf{b} illustrate how the sampling distribution depends on the target.
The sampling distribution is approximately uniform at $\rho^*=0.0941$, whereas several graphs are undersampled at $\rho^*=0.1994$.
The $\chi^2$ tests for uniformity yield $p=0.195$ and $p<10^{-3}$, respectively, confirming this contrast.
These results indicate that DMGG sampling cannot be regarded as uniform across all targets.

Fig.~\ref{fig:small}\textbf{c} places these sampling distributions in the context of ensemble diversity by comparing the exact canonical and microcanonical entropies with the DMGG estimates.
The canonical ensemble maximizes the Shannon entropy at a target mean assortativity, whereas the microcanonical ensemble atains the maximum entropy compatible with an exactly fixed assortativity.
DMGG retains a substantial fraction of the microcanonical Shannon entropy across the feasible range.

To assess whether $S_\text{DI}$ tracks the relative Shannon entropy retained by DMGG, Fig.~\ref{fig:small}\textbf{d} compares $S^\text{DMGG}/S^\text{micro}$ with $S_\text{DI}^\text{DMGG}/S_\text{DI}^\text{micro}$.
The two ratios are strongly correlated (Pearson $r=0.92$).
Although they do not coincide, their agreement in this exactly enumerable system supports the use of $S_\text{DI}$ as a tractable diversity proxy for the larger graphs examined in Fig.~\ref{fig:entropy_efficiency}\textbf{a}, where Shannon entropy estimation is impractical.

Together, the exact small-system analysis and the large-system diagnostics show that DMGG retains substantial configurational diversity.
Combined with its efficiency, this establishes DMGG as a practical method for generating diverse graphs within prescribed macrostates for which microcanonical sampling is computationally inaccessible.

\section{Additional baseline models for DMGG}   \label{app:baseline}
To assess how DMGG combines configurational diversity with efficient generation, we compare it with two complementary baselines.
A greedy heuristic tests the benefit of directing moves toward the target without learning, whereas an adapted ResiNet~\cite{yang2023learning} provides a comparison with an existing RL framework for graph rewiring.

\subsection{Heuristic greedy model}    \label{app:greedy}
Because finding a shortest rewiring trajectory to the target manifold is computationally intractable, we use a greedy baseline that optimizes only the next step.
At each step, the method ranks the attainable changes in assortativity in degree-class space and identifies the valid rewirings that produce the smallest gap $|\rho-\rho^*|$.
If multiple rewirings bring the graph equally close to the target, one of them is selected uniformly at random.
Finding these candidates and sampling uniformly from them can require examining $O(E^2)$ edge pairs in the worst case.
If a move overshoots the target, subsequent steps approach it from the opposite direction.
This procedure is repeated until the target tolerance is reached.

\subsection{Reinforcement learning model: ResiNet} \label{app:resinet}
ResiNet is an inductive RL method for improving network resilience through degree-preserving rewiring.
Whereas DMGG encodes the current graph using Graph Isomorphism Network (GIN) layers with Feature-wise Linear Modulation (FiLM) conditioning, ResiNet uses FireGNN to aggregate representations of the graph together with nested subgraphs obtained by progressively removing high-degree nodes.
This design captures structural responses to targeted attacks but requires processing multiple graph representations at each step.

We adapt ResiNet by conditioning its actor on $\text{sgn}(\rho^*-\rho)$ and its critic on the signed gap $\rho^*-\rho$, as in DMGG.
We use a filtration order of one and the original architectural hyperparameters, while matching the graph families, target range, reward, action masking, and task-dependent PPO rollout and update settings to those of DMGG.
Actions are sampled stochastically from both RL policies during evaluation.

\vspace{\baselineskip}

\begin{figure}
    \centering
    \includegraphics[width=\linewidth]{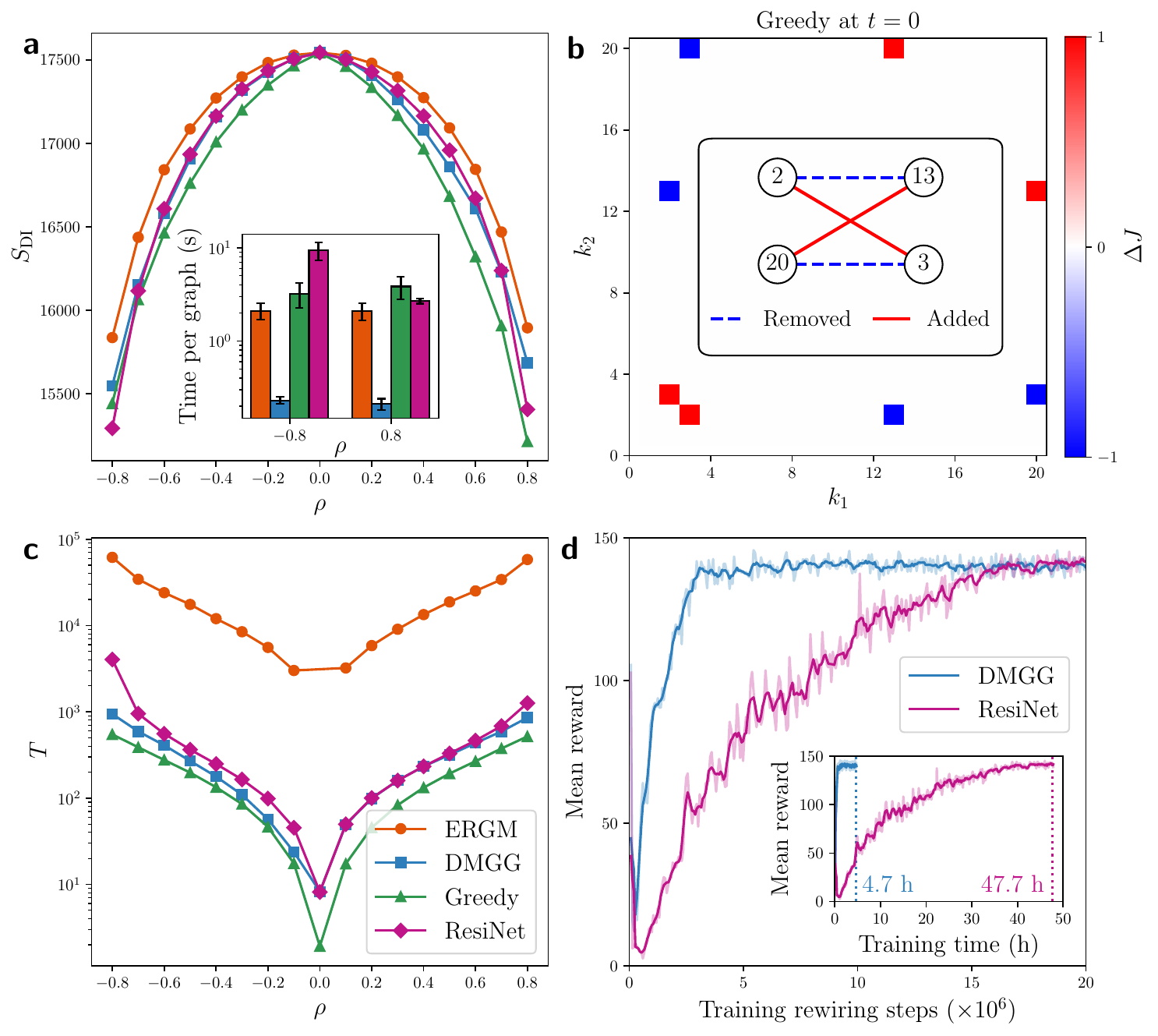}
    \caption{
        \textbf{Additional baseline comparisons}.
        \textbf{a}, Dyad-independent entropy $S_\text{DI}$, shown as in Fig.~\ref{fig:entropy_efficiency}\textbf{a}.
        The inset shows estimated amortized wall-clock time per generated graph in seconds at $\rho=\pm0.8$, with bars and error bars indicating the mean and standard deviation over 1000 graphs.
        \textbf{b}, Joint-degree flux $\Delta J$ of the greedy method at $t=0$ for $\rho^*=0.8$.
        The inset illustrates the corresponding rewiring, with node degrees labeled and removed and added edges shown as blue dashed and red solid lines, respectively.
        \textbf{c}, Required rewirings $T$ versus assortativity $\rho$, shown as in Fig.~\ref{fig:entropy_efficiency}\textbf{c}.
        \textbf{d}, Mean cumulative reward per graph generation during training.
        Faint and solid curves show raw and smoothed values, respectively.
        The inset shows the same reward versus training time, with dotted lines marking the time required for $2 \times 10^7$ training steps.
    } \label{fig:baselines}
\end{figure}

Fig.~\ref{fig:baselines}\textbf{a} compares the diversity of ensembles generated from an ER graph with $N=1000$ and $E=3000$.
The methods exhibit the overall ordering $S_\text{DI}^{\text{Greedy}} < S_\text{DI}^{\text{DMGG}} \simeq S_\text{DI}^{\text{ResiNet}} < S_\text{DI}^{\text{ERGM}}$.
The greedy method minimizes the remaining target gap at each step, sampling randomly only among tied candidates.
Fig.~\ref{fig:baselines}\textbf{b} illustrates the resulting concentration of the joint-degree flux on a few degree pairs.
For the initial state shown, replacing the edges $(2,13)$ and $(3,20)$ with $(2,3)$ and $(13,20)$ yields the largest immediate increase in $K$ among valid rewirings, as illustrated in the inset.
Such restricted choices are consistent with the lower diversity of the greedy ensembles.

Fig.~\ref{fig:baselines}\textbf{c} shows that the greedy method reaches the target with the fewest rewirings $T$, while DMGG and ResiNet require similar values of $T$ over most of the target range.
These results suggest that directing each rewiring toward the target substantially reduces $T$ relative to ERGM.
However, each greedy step can require evaluating $O(E^2)$ candidate rewirings, so $T$ alone does not fully capture its computational cost.

To complement the rewiring counts, the inset in Fig.~\ref{fig:baselines}\textbf{a} reports estimated amortized generation times in seconds at $\rho=\pm0.8$.
The timings reflect practical throughput under different execution conditions: ERGM and the greedy method run on a CPU, whereas DMGG and ResiNet run on a GPU with 32 parallel environments.
Under these conditions, DMGG generates graphs faster than ResiNet, whereas the greedy method takes longer than ERGM despite its smaller $T$.

DMGG is also more efficient to train than ResiNet, as shown in Fig.~\ref{fig:baselines}\textbf{d}.
DMGG approaches its reward plateau within $5\times10^6$ training steps, whereas ResiNet improves more gradually and reaches a comparable reward only by $2\times10^7$ steps.
The inset shows that this budget of $2\times10^7$ training steps takes approximately 4.7 hours for DMGG but 47.7 hours for ResiNet.
ResiNet's greater cost per training step is consistent with FireGNN's additional processing of filtered graph representations, which also adds computational overhead during inference.

To assess when DMGG's training cost is recovered relative to ERGM, we use their estimated amortized generation times of 0.2 and 2.1 seconds per graph at $\rho=\pm0.8$, respectively.
Excluding the cost of tuning $\lambda$, the resulting break-even point under these execution conditions is $4.7\,\mathrm{h}/1.9\,\mathrm{s} \simeq 8.9 \times 10^3$, roughly $10^4$ generated graphs.
Overall, DMGG offers a balanced combination of configurational diversity and efficient generation.

\section*{Data availability}
The data supporting the findings of this study are reproducible using the code described in the Code Availability section.
No additional datasets are required.

\section*{Code availability}
The code used in this study is publicly available at \url{https://github.com/hoyunchoi/DMGG}.

\bibliography{ref}

\section*{Acknowledgements}
This work was supported by the National Research Foundation of Korea (NRF) grant funded by the Korea Government (MSIT) No. RS-2025-00556024 (H. C.) and No. 2022R1A2C1006871 (J. J.), and by a KIAS Individual Grant No. CG079902 (D.-S. L.) at Korea Institute for Advanced Study.
This work is supported by the Center for Advanced Computation at Korea Institute for Advanced Study.

\section*{Author contribution}
H. Y. Choi conceptualized the study, developed the methodology, conducted numerical experiments, analyzed the results, and wrote the original manuscript.
J. Jo and D.-S. Lee supervised the project, provided scientific guidance and feedback throughout the study, and critically reviewed and revised the manuscript.
All authors read and approved the final manuscript.

\section*{Competing interests}
The authors declare no competing interests.

\end{document}